\documentclass{svproc}

\usepackage{url}

\usepackage{amsmath}
\usepackage[numbers]{natbib}
\usepackage{caption}

\usepackage{graphicx}

\newcommand{\pluseq}{\mathrel{+}=}

\begin{document}
\mainmatter              
\title{Use of symmetric kernels for convolutional neural networks}
\titlerunning{Use of symmetric kernels for convolutional neural networks}  
%
\author{Viacheslav Dudar \and Vladimir Semenov}
\authorrunning{Viacheslav Dudar et al.} 
%
\tocauthor{Viacheslav Dudar and Vladimir Semenov}
\institute{Taras Shevchenko National University of Kyiv, Faculty of Computer Science and Cybernetics, Ukraine\\
\email{slavko123@ukr.net, semenov.volodya@gmail.com}}

\maketitle              

\begin{abstract}
At this work we inroduce horizontally symmetric convolutional kernels for CNNs which make the network output invariant to horizontal flips of the image. We also study other types of symmetric kernels which lead to vertical flip invariance, and approximate rotational invariance. We show that usage of such kernels acts as regularizer, and improves generalization of the convolutional neural networks at the cost of more complicated training process.
\end{abstract}

\section{Introduction}
Convolutional neural networks (CNNs) had become one of the most powerful tools in machine learning and computer vision for last last several years \cite{Goodfellow-et-al-2016}. CNNs show state-of-art accuracies for most state-of-art benchmark datasets, such as ImageNet \cite{Krizhevsky12}. CNN has a set of parameters (convolutional kernels, biases, and weights of the last fully connected layers) that are adjusted during the training process. Number of such parameters is typically very large (order of millions or tens of millions). Models with so many parameters do not overfit the data much because of the following reasons:
\begin{itemize}
\item{Data augmentation. Training set is augmented during training in different ways: affine transformations, random subimage selections, random color distortions for each pixel \cite{Krizhevsky12}.}
\item{Efficient regularization techniques. Dropout is one of the most powerful regularization techniques, that corresponds to approximate ensembling over exponential number of subnetworks \cite{Srivastava14}.}
\item{Inner structure of the CNN. Weight sharing is used to enforce approximate invariance of the network output to translations of the input image \cite{Goodfellow-et-al-2016}.}
\end{itemize}
At this work, we focus on CNNs for classification. We propose to make the network output invariant to horizontal image flips via introduction of horizontally symmetric convolutional kernels. Thus we are modifying inner structure of the CNN to enforce additional invariance to improve generalization to the new data. 

\section{Symmetric image kernels}
Let's consider typical CNN architecture that consists of several convolutional layers, followed by elementwise nonlinear function (in most cases it's RELU nonlinearity) alternating with pooling layers (it could be average or max pooling layers) followed by one or several fully connected layers with softmax activation function and trained with categorical cross-entropy loss. 

Consider the first convolutional layer of the net. This layer is translation equivariant, so output of the layer is changed in the same way as the input for translations. But it's not equivariant to the horizontal image flip in case of arbitrary convolution kernel.

We will focus on kernels of size $3\times3$ that are the most widely used \cite{Goodfellow-et-al-2016}. General $3\times3$ convolution kernel:
$$k = \begin{bmatrix}
a & b & c \\
d & e & f \\
g & h & i
\end{bmatrix}
$$
We propose to use horizontally symmetric kernels of the form:
$$k = \begin{bmatrix}
a & b & a \\
d & e & d \\
g & h & g
\end{bmatrix}
$$
We show that in this case convolution layer becomes equivariant to horizontal image flips, and the whole network, under certain structure, becomes invariant to horizontal flips. 

It is enough to show equivariance in one-dimensional case (for each row of the image). 
Consider arbitrary vector:
$$x = \left(x_1\dots x_n\right)$$
and one-dimensional symmetric kernel:
$$k = \left(a,b,a\right)$$
At the moment we consider convolution with stride 1 and no padding.
$$x*k = \left(a\left(x_1+x_3\right)+bx_2\dots a\left(x_{n-2}+x_n\right)+bx_{n-1}\right)$$
Convolution with flipped vector $\hat{x}$:
$$\hat{x}=\left(x_n\dots x_1\right)$$
$$\hat{x}*k = \left(a\left(x_{n-2}+x_n\right)+bx_{n-1}\dots a\left(x_1+x_3\right)+bx_2 \right) = \widehat{x*k}$$
Thus convolution with symmetric kernel of the flipped image is equal to the flip of convolution with initial image. Thus symmetric kernel makes convolution equivariant. Clearly, this result generalizes for 3D convolutions used in CNNs.

Consider now other types of operations performed in CNN. Elementwise application of non-linear function, max and average pooling layers are also clearly flip equivariant.
Thus superposition of 3D convolutions, non-linear functions and poolings is also flip equivariant. 

The only transformation used in CNNs that does not have this property is the flatten layer that maps tensor to vector before fully connected layers. That's why we consider only cases when the last layer is global pooling (max or average). This condition is not restrictive, as the newest architectures (as DenseNet \cite{huang2017densely}) use global pooling before Fully Connected layers. 

Since global pooling (pools tensor to vector of the same depth) is invariant to horizontal flips, the whole network output becomes invariant to horizontal flips. Thus if symmetric kernels are used then posterior probabilities $p\left(C_i|x\right)$ produced by the CNN are exactly the same for the flipped image $\widehat{x}$:
$$p\left(C_i|x\right) = p\left(C_i|\widehat{x}\right)$$

\section{Levels of Symmetry}
We experimented with several levels of symmetry of convolutional kernels. They are summarized in the table 1. Note, that the third column contains induced equivariances for convolutional layers that in turn correspond to induced invariances of the network output (it happens in case global pooling is used before fully connected layer)

\begin{center}
\begin{tabular}{|c|c|c|}
\hline
Symmetry Level & Kernel form & Induced network invariances \\ \hline
0 & $k = \begin{bmatrix}
a & b & c \\
d & e & f \\
g & h & i
\end{bmatrix}$ & No induced invariances \\ \hline
1 & $k =\begin{bmatrix}
a & b & a \\
d & e & d \\
g & h & g
\end{bmatrix}$ & Horizontal flip \\ \hline

2 & $k = \begin{bmatrix}
a & b & a \\
d & e & d \\
a & b & a
\end{bmatrix}$ & \begin{tabular}{@{}c@{}}Horizontal flip \\ Vertical flip\end{tabular} \\ \hline

3 & $k = \begin{bmatrix}
a & b & a \\
b & e & b \\
a & b & a
\end{bmatrix}$ & \begin{tabular}{@{}c@{}}Horizontal flip \\ Vertical flip \\
90 degrees rotations
\end{tabular} \\ \hline
4 & $k = \begin{bmatrix}
a & a & a \\
a & e & a \\
a & a & a
\end{bmatrix}$ & \begin{tabular}{@{}c@{}}Horizontal flip \\ Vertical flip \\
90 degree rotations \\ Approximate arbitrary rotations
\end{tabular} \\ \hline
\end{tabular}
\captionof{table}{Symmetry levels of convolutional kernels}

\end{center}

Different symmetry levels are aplicable to different datasets. For example, for the MNIST dataset levels 2 and higher are not applicable, since one can obtain digit 9 from the digit 6 with consecutive horizontal and vertical flip, so the network trained with such kernels will not distinguish between 6 and 9. But for datasets that contain photos of real world images high symmetry levels are applicable. On the other hand experiments show that training of a network with high symmetry level is a complicated problem, so in practice levels higher that 2 should not be used.

\section {Backpropagation equations}
At this section we describe the modification of the backpropagation procedure that is used to find gradients of the error function with respect to the network weights. For simplicity, we show forward and backward pass of the network only for 1-dimensional convolution for symmetry levels 0 and 1, as extension to 2D convolution and other symmetry levels is straightforward.

Let us denote elements of the convolutional layer in such a way: input vector: $x$, output vector: $y$, general convolutional kernel: $(a,b,c)$, symmetric convolutional kernel: $(a,b,a)$. We denote by $\delta x$ and $\delta y$ derivatives of the error function with respect to vectors $x$ and $y$, and by $\delta a$, $\delta b$, $\delta c$ derivatives of error function with respect to convolutional kernel elements. Equations for forward and backward passes then become:

\begin{center}
\begin{tabular}{|c|c|}
\hline
Level, pass & Operation \\ \hline
Level 0, Forward & $y_i \pluseq ax_{i-1}+bx_{i}+cx_{i+1}$ \\ \hline
Level 1, Forward & $y_i \pluseq a\left(x_{i-1}+x_{i+1}\right) + bx_{i}$ \\ \hline
Level 0, Backward & $\delta x_{i-1} \pluseq a\cdot\delta y_i; \delta x_{i} \pluseq b \cdot\delta y_i; \delta x_{i+1} \pluseq c\cdot\delta y_i$ \\ 
 & $\delta a \pluseq \delta y_i \cdot x_{i-1}; \delta b \pluseq \delta y_i\cdot x_i; \delta c \pluseq \delta y_i\cdot x_{i+1}$\\ \hline
 Level 1, Backward & $\delta x_{i-1} \pluseq a\cdot\delta y_i; \delta x_{i} \pluseq b \cdot\delta y_i; \delta x_{i+1} \pluseq a\cdot\delta y_i$ \\ 
 & $\delta a \pluseq \delta y_i \cdot \left(x_{i-1}+x_{i+1}\right); \delta b \pluseq \delta y_i\cdot x_i$\\ \hline
\end{tabular}
\captionof{table}{Forward and backward pass for symmetric 1D convolution}
\end{center}
Note, that distributive law makes forward and backward pass for level 1 slightly faster than for level 0. The same holds for higher symmetry levels.

\section{Experiments}
To test the given approach, we use CIFAR-10 dataset, which consists of photos of size $3\times32\times 32$ (3 color channels) distributed among 10 classes which include animals, cars, ships and other categories. Training and test sample sizes are 50000 and 10000 correspondingly. As a basic network we chose a variant of DenseNet \cite{huang2017densely} - one of the most efficient recent architectures. Exact configuration of the net we use is given in the table.
\begin{center}
\begin{tabular}{|c|c|}
\hline
Block & Description \\ \hline
Dense block 1 & Number of layers: 1; Convolutional depth: 30 \\
& Input: $3\times32\times32$; Output: $33\times32\times32$ \\ \hline
Pooling 1 & Average pooling $2\times 2$\\
& Output: $33\times16\times16$ \\ \hline
Dense block 2 & Number of layers: 1; Convolutional depth: 30 \\
& Output: $63\times16\times16$ \\ \hline
Pooling 2 & Average pooling $2\times 2$\\
& Output: $63\times8\times8$ \\ \hline
Dense block 3 & Number of layers: 1; Convolutional depth: 30 \\
& Output: $93\times8\times8$ \\ \hline
Pooling 3 & Average pooling $2\times 2$\\
& Output: $93\times4\times4$ \\ \hline
Dense block 4 & Number of layers: 1; Convolutional depth: 30 \\
& Output: $123\times4\times4$ \\ \hline
Pooling 4 & Full Average pooling $4\times 4$\\
& Output: $123\times1\times1$ \\ \hline
Fully Connected & Input length: 123 \\
+ Softmax & Output length: 10  \\ \hline
\end{tabular}
\captionof{table}{CNN configuration}
\end{center}
Note, that we are using RELU nonlinearity for each layer of dense block.

We use this network architecture with each symmetry level for convolutional kernels. Since symmetry levels induce parameter sharing, total number of parameters for next levels is decreased.

We train all the networks with stochastic optimization method ADAM with initial learning rate 0.02, multiplying it by 0.97 after every 5 epochs. We use minibatch size of 1280 in all cases.

Final results for different symmetry levels are given in the table.
\begin{center}
\begin{tabular}{|c|c|c|c|c|c|}
\hline
Level & Model coefficients & Train error & Train accuracy & Test error & Test accuracy \\ \hline
0 & 95520 & 0.19 & 93.15\% & 1.17 & 68.92\% \\ \hline
1 & 62280 & 0.28 & 89.92\% & 1.33 & 69.72\% \\ \hline
2 & 42120 & 0.73 & 74.38\% & 1.07 & 65.54\% \\ \hline
3 & 32040 & 1.02 & 63.75\% & 1.20 & 58.72\% \\ \hline
4 & 21960 & 1.16 & 58.96\% & 1.30 & 54.25\% \\ \hline
\end{tabular}
\captionof{table}{Loss functions and accuracies for different symmetry levels}
\end{center}

\begin{figure}
\includegraphics[width=\textwidth]{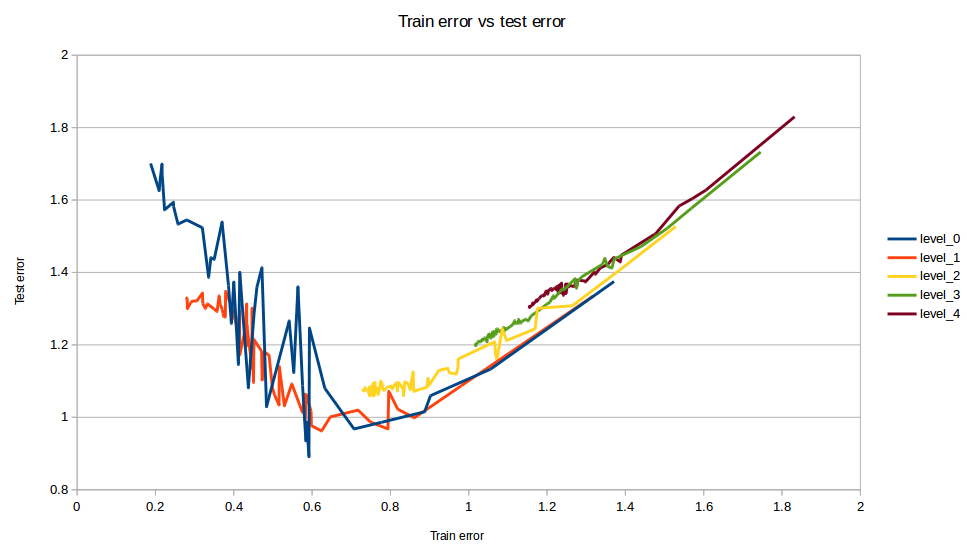}
\caption{Relation between train and test error function values for different symmetry levels}
\label{fig:error}

\includegraphics[width=\textwidth]{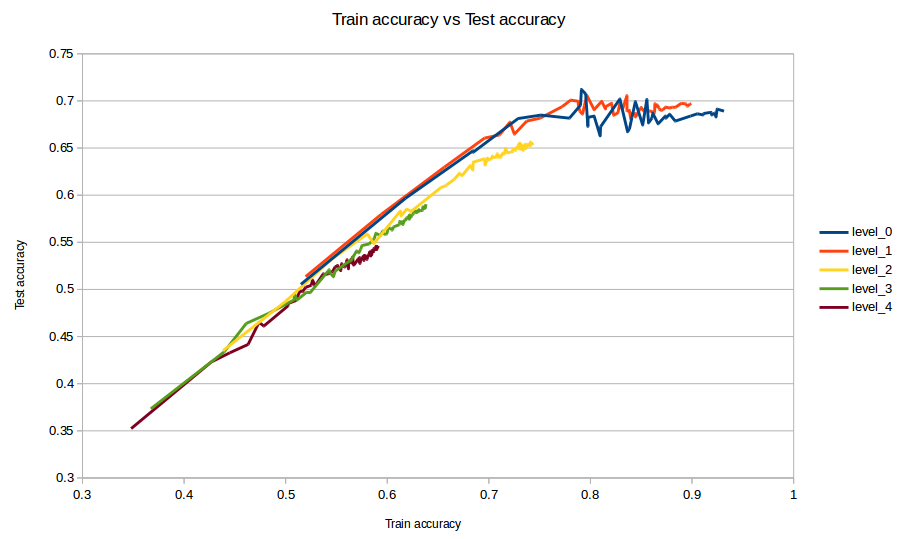}
\caption{Relation between train and test accuracies for different symmetry levels}
\label{fig:accuracy}
\end{figure}

To see if usage of symmeric kernels improves regularization, we recorded train and test error function values and accuracies after every 5-th epoch during training. Scatterplots based on these tables are shown on Figures \ref{fig:error} and \ref{fig:accuracy}. 

\section{Conclusion}
At this work we presented symmetric kernels for convolutional neural networks. Use of such kernels guarantees the network will be invariant under certain transformations, such as horizontal flips for the lowest symmetry level, and approximate rotational symmetry for the highest symmetry level.

We tested this approach by training convolutional neural net with the same DenseNet architecture on CIFAR-10 dataset under different symmetry levels. Since most of the parameters in such network are convolutional kernels (all except biases and $123*10$ matrix for the last fully connected layer) so total number of coefficients adjusted during training varies a lot: from 21960 (highest symmetry level) to 95520 (no symmetry).

Experiments suggest that CNN training is more complicated for higher symmetry levels (as expected) and that only level 1 symmetry shows improvement in generalization. This can be seen on Figure \ref{fig:error} where net without symmetries has higher test error values than net with horizontally symmetric kernels for low train error levels (0.2 - 0.4). The same effect is observed on Figure \ref{fig:accuracy} where the network with horizontally symmetric kernels stabilizes at the highest test accuracy level. This shows networks with horizontally symmetric kernels tend to overfit less.

Why networks with higher symmetry levels (2,3 and 4) do not show improvement in generalization despite providing additional output invariances? From our point of view the reason is as follows. From a common point of view trained convolutional neural network extracts low level image features such as edges and corners at first convolutional layers and combines them into more complex shapes in subsequent layers. With the introduction of DenseNets this interpretation became not so clear since deeper layers have direct dependency on input, but convolutional kernel subtensors acting on input still extract these low level features. The problem with convolutional kernels of high symmetry levels is that they cannot extract image edges or corners of certain orientation (in fact units of convolutional layer respond to edges of different orientations in the same way). Thus such units cannot find joint orientation of edges within the image, besides the general network output is invariant under these transformations. From our point of view this is the reason networks with high symmetry levels do not show improvement in generalization.

Thus we suggest to use convolutional neural networks with horizontally symmetric kernels (symmetry level 1) in practice, since they show lower test error function values and higher test set accuracies as the same network with general convolutional kernels. At the same time such networks have lesser total number of parameters (approximately 2/3) and their output is guaranteed to be invariant under horizontal image flips.

\bibliographystyle{spbasic_unsrt}
\bibliography{refs}

\begin{thebibliography}{6}
\providecommand{\natexlab}[1]{#1}
\providecommand{\url}[1]{{#1}}
\providecommand{\urlprefix}{URL }
\expandafter\ifx\csname urlstyle\endcsname\relax
  \providecommand{\doi}[1]{DOI~\discretionary{}{}{}#1}\else
  \providecommand{\doi}{DOI~\discretionary{}{}{}\begingroup
  \urlstyle{rm}\Url}\fi
\providecommand{\eprint}[2][]{\url{#2}}

\bibitem[{Goodfellow et~al(2016)Goodfellow, Bengio, and
  Courville}]{Goodfellow-et-al-2016}
Goodfellow I, Bengio Y, Courville A (2016) Deep Learning. MIT Press,
  \url{http://www.deeplearningbook.org}

\bibitem[{Krizhevsky et~al(2012)Krizhevsky, Sutskever, and
  Hinton}]{Krizhevsky12}
Krizhevsky A, Sutskever I, Hinton GE (2012) {ImageNet} classification with deep
  convolutional neural networks. In: Proc. Adv. Conf. Neur. Inform. Proc. Syst
  (NIPS 2012), Lake Tahoe, NE

\bibitem[{Srivastava et~al(2014)Srivastava, Hinton, Krizhevsky, Sutskever, and
  Salakhutdinov}]{Srivastava14}
Srivastava N, Hinton G, Krizhevsky A, Sutskever I, Salakhutdinov R (2014)
  Dropout: A simple way to prevent neural networks from overfitting. Journal of
  Machine Learning Research 15:1929--1958

\bibitem[{Huang et~al(2017)Huang, Liu, van~der Maaten, and
  Weinberger}]{huang2017densely}
Huang G, Liu Z, van~der Maaten L, Weinberger KQ (2017) Densely connected
  convolutional networks. In: Proceedings of the IEEE Conference on Computer
  Vision and Pattern Recognition

\bibitem[{He et~al(2015)He, Zhang, Ren, and Sun}]{ResNet}
He K, Zhang X, Ren S, Sun J (2015) Deep residual learning for image
  recognition. CoRR abs/1512.03385,
  \urlprefix\url{http://arxiv.org/abs/1512.03385}, \eprint{1512.03385}

\bibitem[{Kingma and Ba(2014)}]{Kingma14}
Kingma DP, Ba J (2014) Adam: A method for stochastic optimization. In: Int.
  Conf. Learn. Representations, Banff, Canada

\end{thebibliography}

\end{document}